\documentclass[runningheads]{llncs}

\usepackage[mobile]{eccv}

\usepackage{eccvabbrv}
\usepackage{graphicx}
\usepackage[accsupp]{axessibility}  
\usepackage{capt-of,xcolor,xspace,enumitem}
\usepackage{amsmath,amssymb,amsbsy,amsfonts,dsfont,pifont,bm,bbm,mathrsfs,mathtools,nicefrac}
\usepackage{algorithm,algpseudocode,listings}
\usepackage{booktabs,multirow,adjustbox,diagbox,threeparttable,tabularray}
\usepackage[pagebackref,breaklinks,colorlinks,citecolor=eccvblue]{hyperref}
\usepackage{pifont}
\newcommand{\xmark}{\ding{55}}
\newcommand{\cmark}{\ding{51}}

\usepackage{orcidlink}
\usepackage{wrapfig}
\usepackage[capitalize]{cleveref}  
\crefname{section}{Sec.}{Secs.}
\Crefname{section}{Section}{Sections}
\crefname{table}{Tab.}{Tabs.}
\Crefname{table}{Table}{Tables}
\crefname{figure}{Fig.}{Figs.}
\Crefname{figure}{Figure}{Figures}
\crefname{equation}{Eq.}{Eqs.}
\Crefname{equation}{Equation}{Equations}
\newcommand{\method}{\texttt{GRM}\xspace}
\newcommand{\supp}{\textit{Supplementary Material}\xspace}

\usepackage{nicefrac}

\def\gaussians{\mathcal{G}}
\def\gaussian{\mathbf{g}}
\def\R{\mathbb{R}}
\def\mean{\boldsymbol{\mu}}
\def\map{\mathbf{T}}
\def\depth{\tau}
\def\campos{\mathbf{c}_o}
\def\cam{\mathbf{c}}
\def\cams{\mathcal{C}}
\def\ray{\mathbf{r}}
\def\image{\mathbf{I}}
\def\images{\mathcal{I}}
\def\maps{\mathcal{T}}
\def\transformer{E}
\def\feature{\mathbf{F}}
\def\features{\mathcal{F}}
\def\imposenc{\phi}
\def\mask{\mathbf{M}}

\newcommand{\loss}[1]{\mathcal{L}_{\textrm{#1}}}

\DeclareMathOperator{\pixelshuffle}{\textsc{PixelShuffle}}
\DeclareMathOperator{\attention}{\textsc{SelfAttention}}
\DeclareMathOperator{\shift}{\textsc{Shift}}
\DeclareMathOperator{\fullyconnected}{\textsc{Linear}}

\begin{document}

\title{GRM: Large Gaussian Reconstruction Model for \texorpdfstring{\\}~Efficient 3D Reconstruction and Generation}
\titlerunning{\method}

\author{
    Yinghao Xu$^{1}\thanks{Equal Contribution}$ \and
    Zifan Shi$^{1,2\star}$ \and
    Wang Yifan$^{1}$ \and Hansheng Chen$^{1}$ \\ Ceyuan Yang$^{3}$ \and Sida Peng$^{4}$ \and Yujun Shen$^{5}$ \and Gordon Wetzstein$^{1}$
}

\authorrunning{Yinghao Xu et al.}

\institute{
    Stanford University \and 
    The Hong Kong University of Science and Technology \and
    Shanghai AI Laboratory \and 
    Zhejiang University \and
    Ant Group
}

\maketitle

\begin{abstract}
We introduce \method, a large-scale reconstructor capable of recovering a 3D asset from sparse-view images in around 0.1s.
\method is a feed-forward transformer-based model that efficiently incorporates multi-view information to translate the input pixels into pixel-aligned Gaussians, which are unprojected to create a set of densely distributed 3D Gaussians representing a scene.
Together, our transformer architecture and the use of 3D Gaussians unlock a \textit{scalable} and \textit{efficient} reconstruction framework. 
Extensive experimental results demonstrate the superiority of our method over alternatives regarding both reconstruction quality and efficiency.
We also showcase the potential of \method in generative tasks, i.e., text-to-3D and image-to-3D, by integrating it with existing multi-view diffusion models.
Our project website is at: \url{https://justimyhxu.github.io/projects/grm/}. 


\end{abstract}

\begin{figure}[!ht]
    \includegraphics[width=1.0\textwidth]{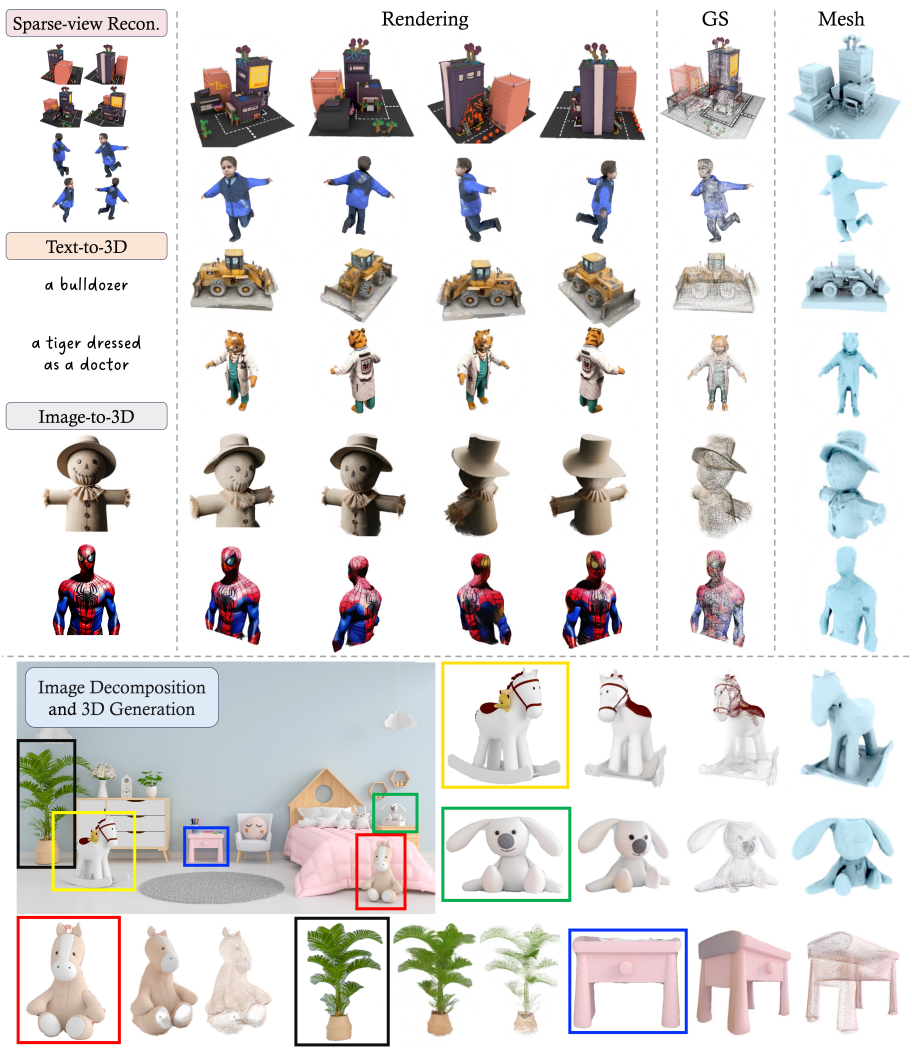}
    \caption{%
        \textbf{High-fidelity 3D assets} produced by \method---a transformer-based reconstruction model built on 3D Gaussians.
        Trained for fast sparse-view reconstruction (top, $\sim$0.1s), \method works in synergy with other tools (e.g., text-to-multiview generation~\cite{li2023instant3d}, image-to-multiview model~\cite{shi2023zero123++}, and 2D segmentation~\cite{kirillov2023segment}) 
        to enable text-to-3D (center top) and image-to-3D (center bottom) generation as well as real-world object reconstruction (bottom).
    }
    \label{figure:teaser}
\end{figure}

\section{Introduction}
\label{sec:intro}

The availability of high-quality and diverse 3D assets is critical in many domains, including robotics, gaming, architecture, among others. Yet, creating these assets has been a tedious manual process, requiring expertise in difficult-to-use computer graphics tools. 

Emerging 3D generative models offer the ability to easily create diverse 3D assets from simple text prompts or single images~\cite{po2023state}. Optimization-based 3D generative methods can produce high-quality assets, but they often require a long time---often hours---to produce a single 3D asset~\cite{poole2022dreamfusion, magic3d, sjc, wang2023prolificdreamer, dreamgaussian}. Recent feed-forward 3D generative methods have demonstrated excellent quality and diversity while offering significant speedups over optimization-based 3D generation approaches~\cite{holodiff,ssdnerf,shen2023gina, renderdiff,viewsetdiff,one2345,hong2023lrm,li2023instant3d,xu2023dmv3d}. These state-of-the-art (SOTA) models, however, typically build on the triplane representation~\cite{eg3d}, which requires inefficient volume rendering. This inefficient rendering step not only hinders fast inference but it often also requires the models to operate at a reduced 3D resolution, limiting representational capacity.

We introduce the Gaussian Reconstruction Model (\method) as a new feed-forward 3D generative model. 
At its core, \method provides a novel sparse-view reconstructor that takes four different views of a 3D object as input and outputs the corresponding 3D scene. 
\method implements two key insights: first, we replace the triplane scene representation of recent feed-forward generative frameworks~\cite{hong2023lrm,li2023instant3d,xu2023dmv3d} by 3D Gaussians; second, we design a pure transformer architecture to translate the set of input pixels to the set of pixel-aligned 3D Gaussians defining the output 3D scene. 
While parts of this architecture uses standard vision transformers (ViT)~\cite{dosovitskiy2020image}, we introduce a new upsampler that utilizes a variation of windowed self-attention layers~\cite{beltagy2020longformer}. 
This upsampler is unique in being able to efficiently pass non-local cues. As demonstrated in our experiment, it is critical for the reconstruction of high-frequency appearance details. 
Instead of attempting to synthesize missing regions from incomplete views---a highly ill-defined problem---we opt to train our model with sparse yet well-distributed views to cover enough information of a scene.
This allows us to allocate the model's capacity for fine-grained detail reconstruction, leading to significantly higher fidelity than relevant baselines for object-level 3D reconstruction. When combined with multi-view diffusion models, \method achieves SOTA quality and speed for text-to-3D and single image-to-3D object generation.

Specifically, our contributions include
\begin{itemize}
\item a novel and efficient feed-forward 3D generative model that builds on 3D Gaussian splatting; 
\item the design of a sparse-view reconstructor using a pure transformer architecture, including encoder and upsampler, for pixel-to-3D Gaussian translation;
\item the demonstration of SOTA quality and speed for object-level sparse-view 3D reconstruction and, when combined with existing multi-view diffusion models, also text-to-3D and image-to-3D generation. 
\end{itemize}

\section{Related Work}

\label{sec:related}

\paragraph{\textbf{Sparse-view Reconstruction.}}
Neural representations, as highlighted in prior works~\cite{occupancy,deepsdf,nerf,sitzmann2019scene,sitzmann2020implicit,chen2022tensorf,mueller2022instant}, present a promising foundation for scene representation and neural rendering~\cite{tewari2022advances}. 
When applied to novel-view synthesis, these methods have demonstrated success in scenarios with multi-view training images, showcasing proficiency in single-scene overfitting. 
Notably, recent advancements~\cite{pixelnerf, mvsnerf, sparseneus, ibrnet, visionnerf, dietnerf} have extended these techniques to operate with a sparse set of views, displaying improved generalization to unseen scenes.
These methods face challenges in capturing multiple modes within large-scale datasets, resulting in a limitation to generate realistic results.
Recent works~\cite{pflrm, hong2023lrm, zou2023triplane} further scale up the model and datasets for better generalization.
But relying on neural volume--based scene representation proves inadequate for efficiently synthesizing high-resolution and high-fidelity images.
Our proposed solution involves the use of pixel-aligned 3D Gaussians~\cite{szymanowicz2023splatter, charatan2023pixelsplat} combined with our effective transformer architecture. 
This approach is designed to elevate both the efficiency and quality of the sparse-view reconstructor when provided with only four input images.

\paragraph{\textbf{3D Generation.}}
The advances of 3D GANs have set the foundation of  3D scene generation. Leveraging various successful 2D GAN architectures~\cite{gan, biggan, PGGAN, StyleGAN, StyleGAN2, StyleGAN3, aurora, gigagan}, 3D GANs~\cite{hologan, graf, pigan, eg3d, giraffe, stylenerf, epigraf, volumegan, discoscene, 3Dsurvey, Get3D, 3dgp} combine 3D scene representations and neural rendering to generate 3D-aware content in a feed-forward fashion. 
Recently, Diffusion Models (DM) have emerged as a more powerful generative model, surpassing GANs in 2D generation~\cite{dhariwal2021diffusion, ddpm, scoresde, ldm}. 
With its extension in 3D being actively explored, we review the most relevant work and refer readers to~\cite{po2023state} for a comprehensive review.
One research line seek to directly train 3D DMs using 3D~\cite{pointe, shape, tridiff, 3dgen, auto3d} or 2D supervision~\cite{holodiff, ssdnerf, shen2023gina, renderdiff, nerfdiff, liu2023zero}. Though impressive, these work either cannot leverage the strong priors from pretrained 2D DMs or they suffer from 3D inconsistency.
Other researchers propose to exploit 2D diffusion priors using an optimization procedure known as Score Distillation Sampling (SDS) and its variant~\cite{poole2022dreamfusion, magic3d, sjc, wang2023prolificdreamer, dreamgaussian, chen2023fantasia3d, hertz2023delta, liu2023zero, shi2023mvdream,chung2023luciddreamer,liang2023luciddreamer}.
These methods yield high-quality 3D generations, but require hours for the optimization process to converge.
Therefore, there is a need to combine the feed-forward generation framework with expressive generative powers from DMs.
To this end, many recent works first use 2D multi-view diffusion and then lift the multi-view inputs to 3D~\cite{genvs, liu2023syncdreamer, tang2023make, one2345, long2023wonder3d,tewari2024diffusion, shi2023zero123++}.
Recently, the Large Reconstruction Model (LRM)~\cite{hong2023lrm} scales up both the model and the dataset to predict a neural radiance field (NeRF) from single-view images. Although LRM is a reconstruction model, it can be combined with DMs to achieve text-to-3D and image-to-3D generation, as demonstrated by extensions such as Instant3D~\cite{li2023instant3d} and DMV3D~\cite{xu2023dmv3d}.
Our method also builds on a strong reconstruction model and uses pretrained 2D DMs to provide input images for 3D generation in a feed-forward fashion. However, we adopt highly efficient 3D Gaussians~\cite{kerbl20233d} for representing and rendering a scene. This design lifts the computation and memory bottleneck posed by NeRF and volumetric rendering, allowing us to to generate high-quality 3D assets within seconds.

Some concurrent works, such as LGM~\cite{tang2024lgm}, AGG~\cite{xu2024agg}, and Splatter Image~\cite{szymanowicz2023splatter}, also use 3D Gaussians in a feed-forward model.
Our model differs from them in the choice of architecture---instead of using conventional convolution-based U-Net, we opt for a purely transformer-based encoder and a highly efficient transformer-based upsampler to generate a large number of pixel-aligned 3D Gaussians, which offering superior reconstruction quality.
\paragraph{\textbf{Generalizable Gaussians.}}
3D Gaussians~\cite{keselman2022approximate,kerbl20233d} and differentiable splatting~\cite{kerbl20233d} have gained broad popularity thanks to their ability to efficiently reconstruct high-fidelity 3D scenes from posed images using only a moderate number of 3D Gaussians.
This representation has been quickly adopted for various application, including image- or text-conditioned 3D~\cite{li2023gaussiandiffusion, chen2023text,liang2023luciddreamer} and 4D generation~\cite{ren2023dreamgaussian4d,ling2023align}, avatar reconstruction~\cite{li2023animatable,zielonka2023drivable,qian2023gaussianavatars,hu2023gaussianavatar,saito2023relightable}, dynamic scene reconstruction~\cite{luiten2023dynamic,wu20234d,yang2023deformable,yang2023real}, among others~\cite{fei20243d,tosi2024nerfs,chen2024survey}.
All of these aforementioned work focus on single-scene optimization, although very recent work also adopts 3D Gaussians into a GAN framework for 3D human generation~\cite{abdal2023gaussian}.

\section{Method}
\label{sec:method}
\method{} is a feed-forward sparse-view 3D reconstructor, utilizing four input images to efficiently infer underlying 3D Gaussians~\cite{kerbl20233d}.
Supplied with a multi-view image generator head~\cite{li2023instant3d, shi2023zero123++}, \method{} can be utilized to generate 3D from text or a single image.
Different from LRM~\cite{hong2023lrm, li2023instant3d, xu2023dmv3d, pflrm}, we leverage \emph{pixel-aligned Gaussians} (\cref{sec:method:gs}) to enhance efficient and reconstruction quality and we adopt a transformer-based network to predict the properties of the Gaussians by associating information from all input views in a memory-efficient way (\cref{sec:method:lgrm}). 
Finally, we detail the training objectives in \cref{sec:method:loss} and demonstrate high-quality text-to-3D and image-to-3D generation in a few seconds (Sec.~\ref{sec:method:gen}).

\subsection{Pixel-aligned Gaussians}\label{sec:method:gs}
3D Gaussians use a sparse set of primitives $\gaussians = \{\gaussian_i\}_{i=1}^{N}$ to represent the geometry and appearance of a 3D scene,
where each Gaussian is parameterized with location $\mean\in\R^3$, rotation quaternion $\mathbf{r}\in\R^4$, scale $\mathbf{s}\in\R^3$, opacity $o\in\R$, and the spherical harmonics (SH) coefficients $\mathbf{c}\in\R^D$, with $D$ denoting the number of SH bases.
These Gaussians can be rendered in real time using the differentiable rasterizer~\cite{kerbl20233d}.
3D Gaussians have gained tremendous popularity for single-scene optimization (SSO), but utilizing them in a generalizable framework remains challenging. 
A primary reason is that the properties of Gaussians are highly inter-dependent---multiple configurations can lead to the same visual representation, causing optimization difficulty. 
On the other hand, 3D Gaussians are an unstructured 3D representation, making it challenging to integrate effectively with neural networks for predictions which potentially increases the complexity of the prediction process.
We introduce pixel-aligned Gaussians~\cite{charatan2023pixelsplat, szymanowicz2023splatter} to address the above challenges. Instead of directly predicting a set of Gaussians and expect them to accurately cover the entire shape, we constrain the Gaussians' locations along the input viewing rays, \ie, 
\begin{equation}
    \mean = \campos + \depth \ray, 
\end{equation}
where \(\campos\) and \(\ray\) denote the camera center and the ray direction.
Specifically, for every input view we predict a Gaussian attribute map \(\map\in\R^{H\times W\times C}\) of \(C=12\) channels, corresponding to depth $\depth$, rotation, scaling, opacity, and the DC term of the SH coefficients.
We then unproject the pixel-aligned Gaussians into 3D, producing a total of $V\times H\times W$ densely distributed 3D Gaussians.
Pixel-aligned Gaussians establish connections between input pixels and the 3D space in a more direct way, alleviating the learning difficulty, resulting to better reconstruction quality as we empirically show in \cref{sec:exp:ablation}.

\begin{figure}[t!]
    \centering
	\includegraphics[width=1\linewidth]{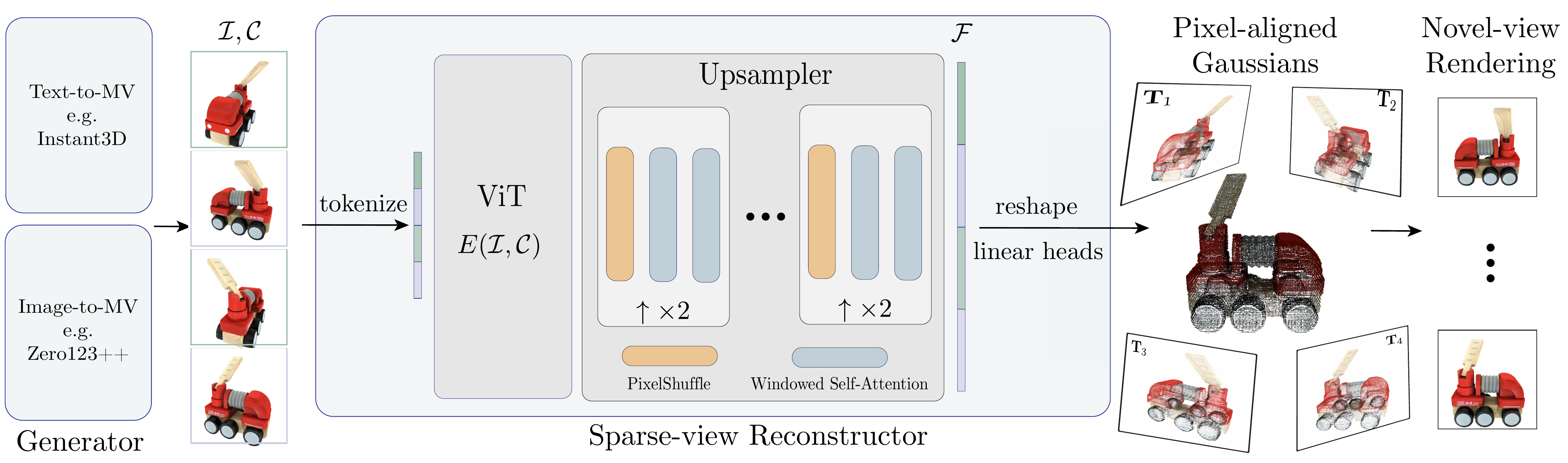}
	\caption{\textbf{\method pipeline}. Given 4 input views, which can be generated from text~\cite{li2023instant3d} or a single image~\cite{shi2023zero123++}, our sparse-view reconstructor estimates the underlying 3D scene in a single feed-forward pass using pixel-aligned Gaussians. The transformer-based sparse-view reconstructor, equipped with a novel transformer-based upsampler, is capable of leveraging long-range visual cues to efficiently generate a large number of 3D Gaussians for high-fidelity 3D reconstruction.}
	\label{fig:pipeline}
\end{figure}

\subsection{Large Gaussian Reconstruction Model}\label{sec:method:lgrm}
In the following, we introduce our network, which transforms a set of input images $\images = \{\image_v\}_{v=1}^{V}$ and their camera poses \(\cams=\{\cam_v\}_{v=1}^{V}\) to the Gaussian maps $\maps = \{\map_v\}_{v=1}^{V}$.

\paragraph{\textbf{Transformer-based Encoder.}}
For a given input image \(\image_v\in\R^{H\times W\times 3}\), we first inject the camera information to every pixel following~\cite{xu2023dmv3d, lfn} with Pl\"{u}cker embedding~\cite{jia2020plucker, lfn}.
Then we use a convolutional image tokenizer with kernel and stride 16 to extract local image features, resulting in a \(\nicefrac{H}{16} \times \nicefrac{W}{16}\) feature map.
The features from every view are concatenated together to a single feature vector of length \(\left(V\times \nicefrac{H}{16}\times \nicefrac{W}{16}\right)\).
Following common practice in vision transformers, we append learnable image position encodings for each token to encode the spatial information in the image space.
The resulting feature vector is subsequently fed to a series of self-attention layers.
The self-attention layers attend to all tokens across all the input views, ensuring mutual information exchange among all input views, resembling traditional feature matching and encouraging consistent predictions for pixels belonging to different images.
The output of the transformer-based encoder is a \(V\times \nicefrac{H}{16} \times \nicefrac{W}{16}\)-long feature vector, denoted as \(\feature\).
Formally, the transformer function can be written as
\begin{align}
\feature = \transformer_{\theta, \imposenc}\left(\mathcal{I}, \mathcal{C}\right),
\end{align}
where $\theta$ and $\imposenc$ denote the network parameters and the learnable image position encodings.

In the transformer encoder, we utilize patch convolution to tokenize the input images, resulting in the output feature $\feature$ with a smaller spatial dimension.
While this is advantageous for capturing broader image context, it is limited in modeling high-frequency details.
To this end, we introduce a transformer-based upsampler to improve the detail reconstruction.

\paragraph{\textbf{Transformer-based Upsampler.}}
Inspired by previous work~\cite{beltagy2020longformer,liu2021swin}, we use  windowed attention to balance the need for non-local multi-view information aggregation and feasible computation cost.
Specifically, we construct multiple upsampler blocks to progressively upsample the features by factors of 2 until we reach the original input image resolution. 
In each block, we first quadruple the feature dimensions with a linear layer and then double the spatial dimension with a PixelShuffle layer~\cite{shi2016real}.
The upsampled feature maps are grouped and passed to a self-attention layer in a sliding window of size \(W\) and shift \(\nicefrac{W}{2}\).
While the self-attention operation is performed within each distinct window, to maintain manageable memory and computation, the overlapping between shifted windows improves non-local information flow.
Formally, an upsampler block contains the following operations:
\begin{align}
\feature &= \pixelshuffle\left(\fullyconnected\left(\feature\right), 2\right), \\
\feature &= \attention\left(\feature, W\right), \\
\feature &= \shift\left(\attention\left(\shift\left(\feature, W/2\right), W\right), -W/2\right).
\end{align}
After several blocks, the context length expands to the same spatial dimension as the input.
We reshape the features back to 2D tensors, resulting in $V$ feature maps with a resolution of ${H}\times {W}$, denoted as \(\features = \{\feature_v\}_{v=1}^{V}\).

\paragraph{\textbf{Rendering with Gaussian Splatting.}}
From the upsampled features \(\features\), we predict the Gaussian attribute maps \(\{\map_v\}_{v=1}^{V}\) for pixel-aligned Gaussians using separate linear heads. 
As mentioned in ~\cref{sec:method:gs}, these are then unprojected along the viewing ray according to the predicted depth, from which a final image \(\image_{v'}\) and alpha mask \(\mask_{v'}\) (used for supervision) can be rendered at an arbitrary camera view \(\cam_{v'}\) through Gaussian splatting. 

\subsection{Training}\label{sec:method:loss}
During the training phase, we sample $V=4$ input views that sufficiently cover the whole scene, and supervise with additional views to guide the reconstruction.
To remove floaters, we also supervise the alpha map from Gaussian splatting with the ground truth object mask available from the training data.

Given $V'$ supervision views, the training objective is 
\begin{align}
    \loss{} &= \dfrac{1}{V'}\sum_{{1 \leq v'\leq V'}} \loss{img}  + \loss{mask},\\
    \loss{img} &= L_2\left(\image_{v'}, \hat{\image}_{v'}\right) + 0.5 L_p\left(\image_{v'}, \hat{\image}_{v'}\right),\\
    \loss{mask} &= L_2\left(\mask_{v'}, \hat{\mask}_{v'}\right),
\end{align}
where $\hat{\image}_{v'}$ and $\hat{\mask}_{v'}$ denote ground truth image and alpha mask, respectively. $L_2$ and $L_p$ are L2 loss and perceptual loss~\cite{johnson2016perceptual}.

To further constrain the Gaussian scaling, we employ the following activation function corresponding to the output $\mathbf{s}_o$ of the linear head for scale. Subsequently, we conduct linear interpolation within predefined scale values $s_{min}$ and $s_{max}$:
\begin{align}
    \mathbf{s} = s_{min}\textsc{Sigmoid}(\mathbf{s}_o) + s_{max}(1- \textsc{Sigmoid}(\mathbf{s}_o)).\label{eq:scaling_act}
\end{align}

\subsection{Reconstructor for 3D Generation}\label{sec:method:gen}
Our reconstructor alone is able to efficiently estimate a 3D scene from 4 input images. 
We can seamlessly integrate this reconstructor with any diffusion model that generates multi-view images to enable fast text-to-3D and image-to-3D generation, similar to Instant3D~\cite{li2023instant3d}.
Specifically, we use the first stages of Instant3D~\cite{li2023instant3d} and Zero123++\cite{shi2023zero123++} to produce 4 multi-view images from a text prompt or a single image, respectively. Note that Zero123++ generates 6 images, from which we select the 1$^{st}$, 3$^{rd}$, 5$^{th}$, and 6$^{th}$ as input to our reconstructor.

\section{Experiments}
\label{sec:exp}

\subsection{Experimental Settings}

\paragraph{\textbf{Training Settings.}}
The encoder \(E\) consists of 1 strided convolution layer to tokenize the image and 24 self-attention layers with channel width 768.
The upsampler consists of 4 upsampler blocks and each block contains 2 attention layers.
For training, we use AdamW~\cite{loshchilov2017decoupled} with a learning rate initialized at $0.0003$ decayed with cosine annealing after 3$k$ steps.
Deferred back-propagation~\cite{zhang2022arf} is adopted to optimize GPU memory.
We train our model on 32 NVIDIA A100 GPUs for 40M images on the resolution of 512$\times$512, using a batch size of 8 per GPU and taking about 4 days to complete.
The window size in the transformer-upsampler is 4096.
The values for $s_{min}$ and $s_{max}$ are set to 0.005 and 0.02.

\paragraph{\textbf{Training Data.}}
We obtain multi-view images from Objaverse~\cite{objaverse} as training inputs.
Objaverse contains more than 800$k$ 3D objects with varied quality.
Following \cite{ li2023instant3d}, we filter 100$k$ high-quality objects, and render 32 images at random viewpoints with a fixed 50$^\circ$ field of view under ambient lighting.

\paragraph{\textbf{Test Data.}}
We use Google Scanned Objects (GSO)~\cite{gso}, and render a total of 64 test views with equidistant azimuth at $\{10, 20, 30, 40\}$ degree elevations.
In sparse-view reconstruction, the evaluation uses full renderings from 100 objects to assess all models. 
For single-view reconstruction, we restrict the analysis to renderings generated at an elevation angle of 20 from 250 objects.
More details about training settings and data are presented in \supp.

\begin{figure}[th!] 
	\centering
	\includegraphics[width=\linewidth]{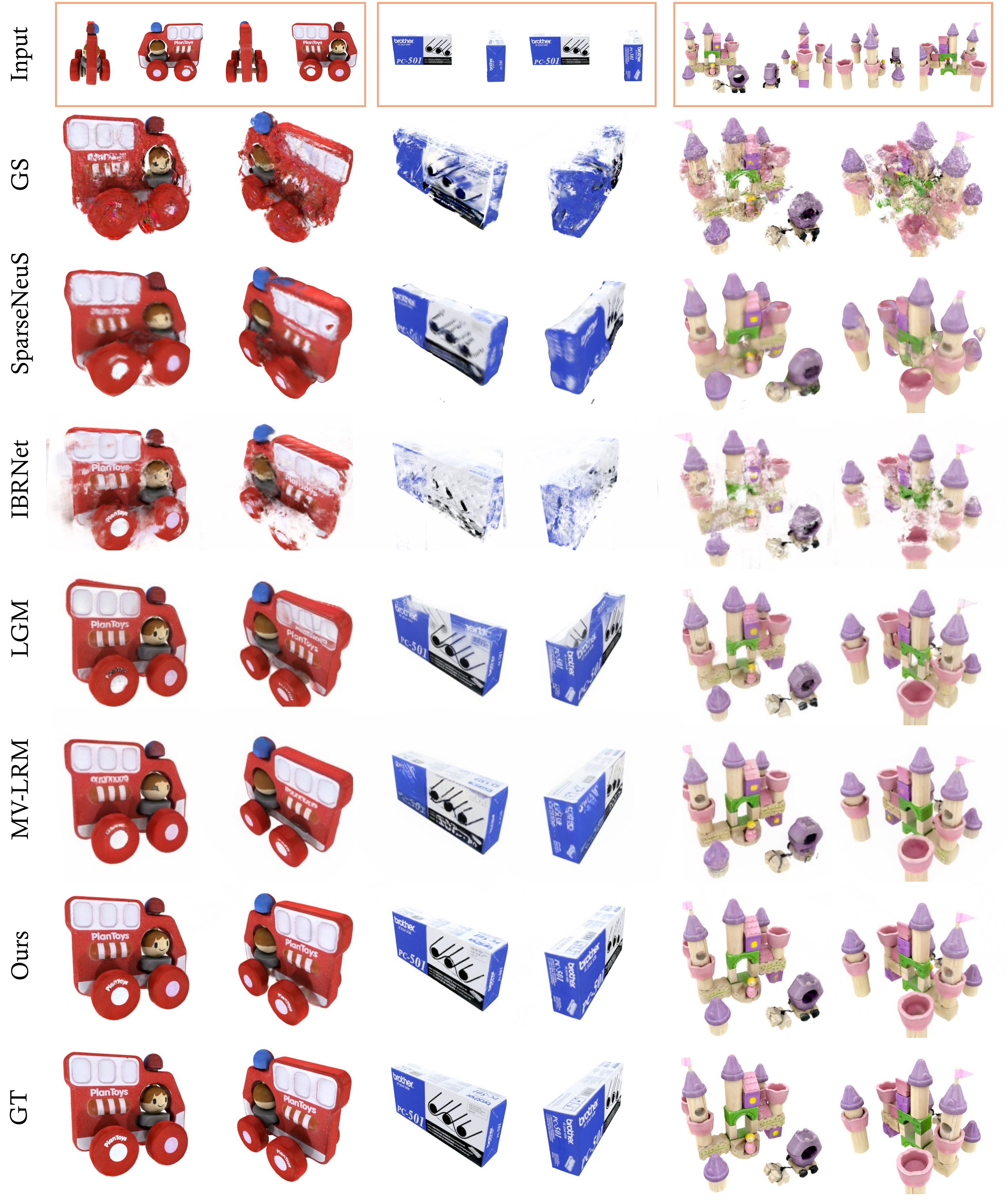}
	\caption{\textbf{Sparse-view reconstruction.} Given the same sparse-view inputs (top), we compare the 3D reconstruction quality with strong baselines, among which GS~\cite{kerbl20233d} is optimization based. SparseNeuS~\cite{sparseneus} (trained in One-2-3-45~\cite{one2345}) and IBRNet~\cite{ibrnet} require 16 views (only 4 of those are visualized in the top row). \method{} more accurately reconstructs the geometric structure as well as the finer appearance details.}
	\label{fig:reconstruction}
\end{figure}

\subsection{Sparse-view Reconstruction}

\paragraph{\textbf{Baselines and Metrics.}}
We compare our method with Gaussian Splatting~\cite{kerbl20233d}, SparseNeuS~\cite{one2345, sparseneus}, IBRNet~\cite{ibrnet}, MV-LRM~\cite{li2023instant3d}, and the concurrent LGM~\cite{tang2024lgm}. 
Since MV-LRM neither released their code nor model, we reproduce it following their paper. The original SparseNeuS does not support 360$^\circ$ reconstruction, so we use the improved version trained in One-2-3-45~\cite{one2345}. The remaining baselines are evaluated using their original implementations. 
For all baselines, except for SparseNeuS and IBRNet, we use the same set of four input views that roughly cover the entire object. 
SparseNeuS and IBRNet are originally intended for denser input views, hence we additionally sample another $12$ views, resulting in 16 roughly uniformly distributed views.
Following MV-LRM, we use PSNR, SSIM, and LPIPS to measure the reconstruction quality. 
Additional evaluations for geometry reconstruction accuracy is included in the \supp.
We conduct all the comparisons using a resolution of 512$\times$512.

\begin{table}[h]

\centering
\scriptsize
\caption{\textbf{Sparse-view reconstruction.} We compare the reconstruction quality for 64 novel views with 100 objects from GSO~\cite{gso}. \method yields superior quality while maintaining fast speed. INF. Time is time from image inputs to the corresponding 3D representation (e.g., 3D Gaussians or Triplane NeRF); REND. Time is the time used to render a $512^2$ image from the 3D representation.}
\label{tab:reconstruction}

\begin{tabular}{lcccc|cc}
\toprule
Method & \#views & PSNR\textuparrow & SSIM\textuparrow & LPIPS\textdownarrow & INF. Time\textdownarrow & REND. Time\textdownarrow \\
\hline

GS~\cite{kerbl20233d}  & 4 & 21.22 & 0.854 & 0.140 & 9 min & Real time\\

IBRNet~\cite{ibrnet}    & 16 & 21.50 & 0.877 & 0.155 & 21 sec & 1.2 sec\\
SparseNeuS~\cite{one2345, sparseneus} & 16 & 22.60 & 0.873 & 0.132 & 6 sec & Real time\\
LGM~\cite{tang2024lgm}             & 4 & 23.79 & 0.882 & 0.097 & 0.07 sec & Real time\\
MV-LRM~\cite{li2023instant3d}             & 4 & 25.38 & 0.897 & 0.068 & 0.25 sec & 1.7 sec \\
\method{} (Ours) & 4 & \bf{30.05} & \bf{0.906} & \bf{0.052} & 0.11 sec & Real time\\
\bottomrule
\end{tabular}
\vspace{-5pt}
\end{table}

\paragraph{\textbf{Results.}}
As \cref{tab:reconstruction} shows,
our method significantly outperforms all baselines across all metrics by a large margin, even though SparseNeuS and IBRNet require 4 times more input views.
At the same time, our inference speed is among the fastest two, only second to the concurrent LGM.
However, ours predicts 16 times more Gaussians than they do, which leads to a much higher reconstruction fidelity.
\cref{fig:reconstruction} shows the novel-view rendering results.
Compared to other methods, our reconstructions accurately reflect the geometric structures, containing no visible floaters, and they capture better appearance details than the baselines.

\subsection{Single Image-to-3D Generation}\label{sec:exp:image3D}
As shown in \cref{sec:method:gen}, \method can be used for single image-to-3D generation by combining it with image-to-multiview diffusion models, such as Zero123++\cite{shi2023zero123++}. 

\paragraph{\textbf{Baselines and Metrics.}}
The baselines include SOTA single-image 3D generation methods: ShapE~\cite{shape}, One-2-3-45~\cite{one2345}, One-2-3-45++~\cite{liu2023one}, DreamGaussian~\cite{dreamgaussian}, Wonder3D~\cite{long2023wonder3d}, TriplaneGaussian~\cite{zou2023triplane}, and LGM~\cite{tang2024lgm}.
For all methods, we use the same input image that is selected randomly from the 4 input views in the sparse-view reconstruction task.
All the comparisons are done using a resolution of 256$\times$256.

\begin{figure}[t!]
\centering
\includegraphics[width=1\linewidth]{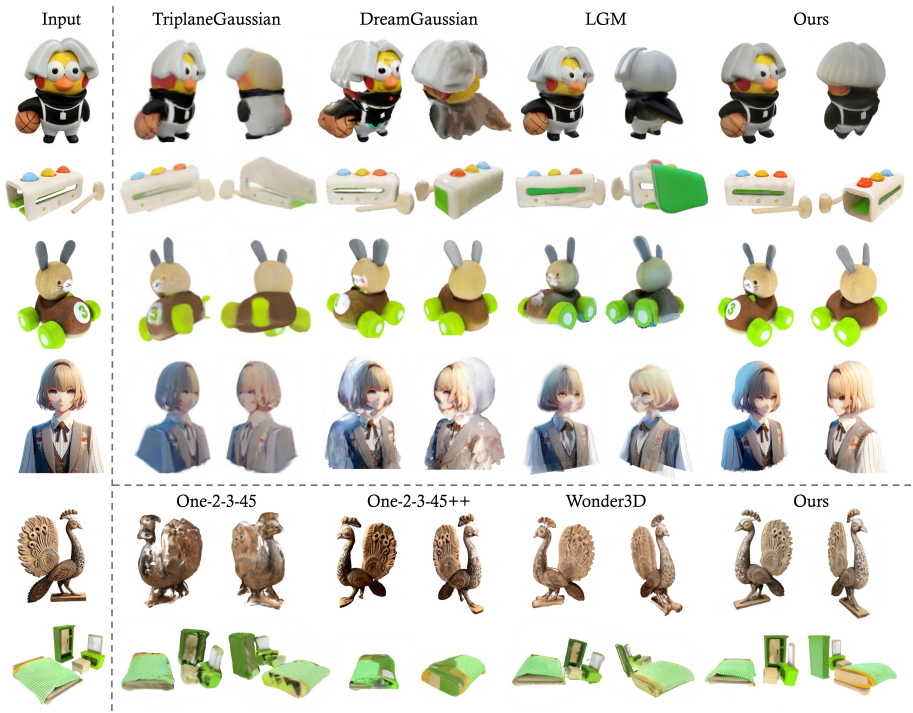}
\caption{\textbf{Single image-to-3D generation.} We compare with methods using Gaussians (top) and non-Gaussians (bottom) as 3D representations. Reconstruction methods, e.g., TriplaneGaussian~\cite{zou2023triplane} struggle to realistically complete the unseen region (rows 1--2). SDS-based methods, e.g., DreamGaussian~\cite{dreamgaussian} suffer from considerable inconsistencies with the input image. LGM~\cite{tang2024lgm}, One-2-3-45~\cite{one2345}, One-2-3-45++~\cite{liu2023one}, and Wonder3D~\cite{long2023wonder3d} also combine multiview diffusion and reconstruction for single image-to-3D generation, but they produce blurrier texture and noticeable geometry artifacts. Our results contain more appearance details and shows significantly better consistency with the input.}
	\label{fig:image-to-3d}
 \vspace{-5pt}
\end{figure}

Similar to sparse-view reconstruction, we compute PSNR, SSIM and LPIPS. 
We also include CLIP scores~\cite{radford2021learning} and FID~\cite{heusel2017gans}, which are two common metrics to evaluate image similarity in generation tasks~\cite{liu2023zero,one2345, xu2023dmv3d}.
Geometry evaluations are included in the \supp.

\begin{table}[thb!]
\centering
\scriptsize
\caption{\textbf{Single image-to-3D generation.} Combined with an image-to-multiview diffusion model~\cite{shi2023zero123++}, \method{} can be used for single image-to-3D generation. Our method outperforms relevant baselines in terms of the quality of the synthesized novel views with fast inference speed. }
\label{tab:image-to-3d}

\begin{tabular}{lccccccc}
\toprule
Method              & PSNR\textuparrow & SSIM\textuparrow & LPIPS\textdownarrow & CLIP\textuparrow & FID\textdownarrow  & INF. Time\textdownarrow \\ \midrule
One-2-3-45~\cite{one2345}             & 17.84 & 0.800 & 0.199 & 0.832 & 89.4  & 45 sec  \\
Shap-E~\cite{shape}               & 15.45  & 0.772 & 0.297 & 0.854 & 56.5 & 9 sec  \\
DreamGaussian~\cite{dreamgaussian}       &  19.19 & 0.811 & 0.171 & 0.862 & 57.6 & 2 min \\
Wonder3D~\cite{long2023wonder3d}            & 17.29 & 0.815 & 0.240 & 0.871 & 55.7  & 3 min \\
One-2-3-45++~\cite{liu2023one}           & 17.79 & 0.819 & 0.219 & 0.886 & 42.1 & 1 min \\

\midrule
TriplaneGaussian~\cite{zou2023triplane}    & 16.81 & 0.797 & 0.257 & 0.840 & 52.6  & 0.2 sec \\ 
LGM ~\cite{tang2024lgm}                & 16.90 & 0.819 & 0.235 & 0.855 & 42.1  & 5 sec\\
\method{} (Ours)    & \textbf{20.10}  & \textbf{0.826} & \textbf{0.136} & \textbf{0.932} & \textbf{27.4} & 5 sec \\

\bottomrule
\end{tabular}
\vspace{-5pt}
\end{table}

\paragraph{\textbf{Results.}}
The quantitative results are presented in \cref{tab:image-to-3d}. Notably, \method outperforms all baselines across all metrics. 
Our model only takes 5 seconds in total to generate 3D Gaussians from the input image, which includes the runtime of the generation head. While this is slower than TriplaneGaussian, we achieve significantly higher reconstruction quality.
Our advantage is further demonstrated in the qualitative results shown in \cref{fig:image-to-3d}. On the top, we compare with other 3D Gaussian-based methods. The pure reconstruction method TriplaneGaussian struggles to fill in the missing content realistically (see rows 1--2). DreamGaussian, using SDS optimization, shows various geometry artifacts (row 1) and overall noticeable inconsistencies with the input image. LGM, also using an image-to-MV generation head, produces blurrier and inconsistent texture and geometry. 

The bottom of \cref{fig:image-to-3d} shows non-Gaussians based approaches. These methods all display various geometry and appearance artifacts, inconsistent with the input.
In contrast, our scalable \method learns robust data priors from extensive training data, demonstrating strong generalization ability on generated multi-view input images with accurate geometry and sharper details. This leads to fast 3D generation and state-of-the-art single-image 3D reconstruction.

\subsection{Text-to-3D Generation}
By using a text-to-MV diffusion model, such as the first stage of Instant3D~\cite{li2023instant3d}, \method can generate 3D assets from text prompts.

\paragraph{\textbf{Baselines and metrics.}}
We choose Shap-E~\cite{shape}, Instant3D~\cite{li2023instant3d}, LGM~\cite{tang2024lgm}, and MVDream~\cite{shi2023mvdream} as baselines. MVDream represents the SOTA of optimization-based methods, while others are feed-forward methods.
We use the 200 text prompts from DreamFusion~\cite{poole2022dreamfusion}. The metrics we use are CLIP Precisions~\cite{dreamfields, dreamgaussian}, Averaged Precision~\cite{xu2023dmv3d}, CLIP Score~\cite{li2023instant3d, xu2023dmv3d}, which measure the alignment between the text and images.
All the comparisons are done using a resolution of 512$\times$512.
Additionally, we include a preference study on Amazon Mechanical Turk, where we recruited 90 unique users to compare the generations for 50 text prompts.

\begin{figure}[t]
	\centering
	\includegraphics[width=1.0\linewidth]{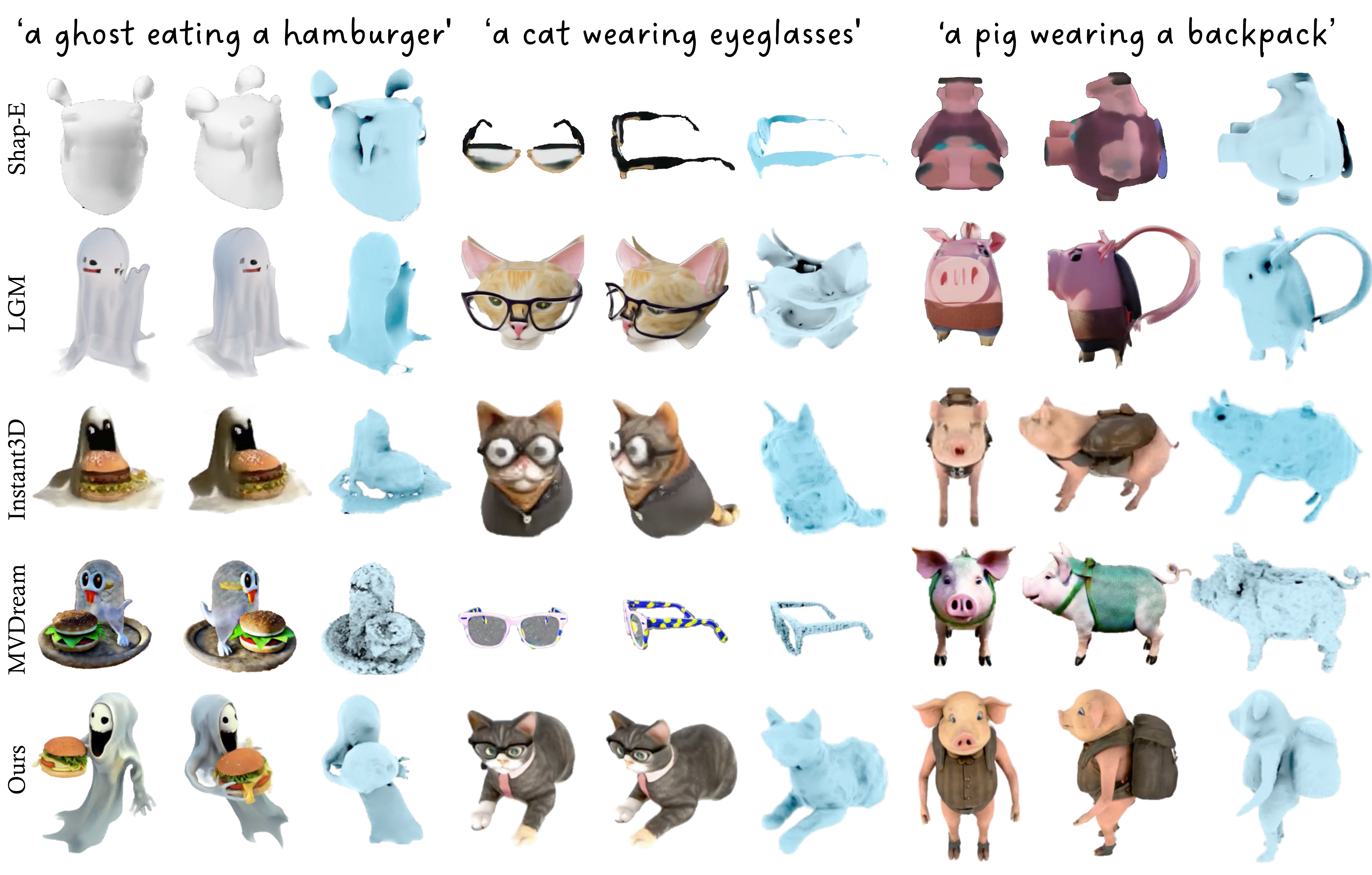}
	\caption{\textbf{Text-to-3D Generation.} Our method creates high-quality 3D assets from text prompts with accurate text alignment. \method only requires 8 seconds to generate comparable results to the SOTA optimization-based MVDream, which takes 1 hour.}
	\label{fig:text-to-3d}
\end{figure}

\paragraph{\textbf{Results.}}
As shown in \cref{tab:text-to-3d}, our method consistently ranks the highest among feed-forward methods (rows 1--3) and compares onpar with optimization-based MVDream. Visually, as shown in \cref{fig:text-to-3d}, our method excels at generating plausible geometry and highly detailed texture.
MVDream, using SDS-based optimization, requires 1 hours to generate a single scene. 
It delivers impressive visual quality, but exhibits sub-optimal text-image alignment, as indicated by the CLIP score and the `a cat wearing eyeglasses' example in \cref{fig:image-to-3d}.

\begin{table}[thb!]
\centering
\scriptsize
\caption{\textbf{Text-to-3D Generation}. Combined with a text-to-multiview diffusion model~\cite{li2023instant3d}, \method{} can be used for text-to-3D generation and it achieves a competitive CLIP score. Our method is most often preferred in our user study while maintains very fast inference speed.}
\begin{tabular}{lccccc}
\toprule
Method & R-Prec\textuparrow & AP\textuparrow & CLIP\textuparrow & User Pref\textuparrow & INF. Time \\ \midrule
Shap-E~\cite{shape}  & 12.7 & 17.7 & 17.3 & 15.7\%  & 9 sec\\
LGM~\cite{tang2024lgm}        & 35.8 & 41.4 & 17.2 & 13.3\% & 5 sec \\
Instant3D~\cite{li2023instant3d} & 59.3& 64.3 &17.9 & 15.7\% & 20 sec\\
MVDream-SDS~\cite{shi2023mvdream} &  \textbf{70.1} & \textbf{74.4} & 17.6 & 25.9\% & 1 hour\\
\method{} (ours) & 67.5 & 72.0 & \textbf{18.5} & \textbf{29.5}\% & 8 sec \\
\bottomrule
\end{tabular}
\label{tab:text-to-3d}
\vspace{-20pt}
\end{table}

\begin{table}[h]
\centering
\caption{\textbf{Ablation.} \textbf{Left:} 
Using the \textsc{sigmoid} activation improves the visual quality across all metrics; increasing the number of sampling blocks also increases the Gaussians' density and their modeling capability, as demonstrated by the growing trend of PSNR; finally supervising the alpha channel further boost the reconstruction quality by removing outlier Gaussians. 
\textbf{Right:} We ablate the proposed transformer-based upsampler and pixel-aligned Gaussians using alternative approaches, and demonstrate that each component is critical for the final reconstruction quality.}
\begin{subtable}[b]{0.4\linewidth}
\resizebox{\columnwidth}{!}{%

\begin{tabular}{ccc|ccc}
\toprule
Scale Act & \#Up & $\alpha$-reg & PSNR & SSIM & LPIPS \\\midrule
\xmark & 0 & \xmark & 24.43 & 0.638 & 0.133\\ \hline
\cmark & 0 & \xmark  &  27.51 & 0.900 & 0.044  \\ \hline
\cmark & 1 & \xmark & 29.11 & 0.922 & 0.037\\
\cmark & 3 & \xmark & 29.38 &0.917 & 0.036 \\ \hline
\cmark & 3 & \cmark  &  29.48 & 0.920 & 0.031\\
\bottomrule
\end{tabular}}
\end{subtable}\hfill
\begin{subtable}[b]{0.45\linewidth}

\resizebox{\columnwidth}{!}{%
\begin{tabular}{p{0.52\linewidth}ccc}
\toprule
Method & PSNR & SSIM & LPIPS  \\\midrule
Conv-Upsampler & 27.23 & 0.894 & 0.063 \\
XYZ prediction & 28.61 & 0.910 & 0.037 \\ \hline
Full model  & 29.48 & 0.920 & 0.031\\
\bottomrule
\end{tabular}}
\end{subtable}
\label{tab:abla}
\vspace{-10pt}
\end{table}

\subsection{Ablation Study}\label{sec:exp:ablation}
We analyze our model components and architectural design choices on the training resolution of 256. 
The results are shown in \cref{tab:abla}.
Note that all ablations are trained with 16 GPUs for 14M images.

\paragraph{\textbf{Scale Activation.}}
We conduct an experiment to study the function of the activation function for the Gaussians' scales.
Conventionally, gaussian splatting~\cite{kerbl20233d} takes the scales with an exponential activation.
However, the exponential activation can easily result in very large Gaussians, resulting in unstable training and blurry images.
With the linear interpolation between the predefined scale ranges, the model achieves better appearance regarding all metrics in the first two rows of \cref{tab:abla}.

\paragraph{\textbf{Number of Upsampler Blocks.}}
We analyze the effects of different number of upsampler blocks.
We can see that the model performance increases as the number of upsampler blocks grows from 0 to 3 (\cref{tab:abla}, left, rows 2--4), benefiting from the detailed spatial features modulated by our transformer-based upsampler.

\begin{wrapfigure}{r}{0.3\textwidth}
  \begin{center}
    \includegraphics[width=0.28\textwidth]{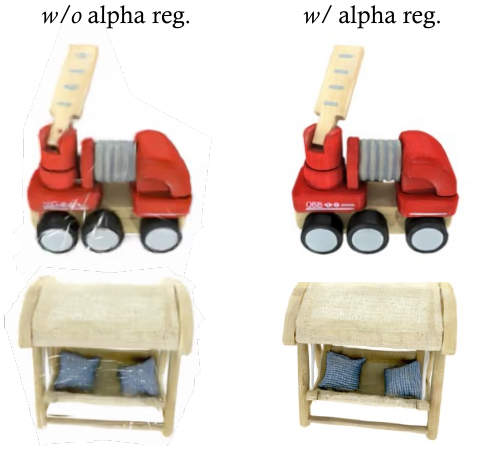}
  \end{center}
  \vspace{-10pt}
  \caption{Comparison on alpha regularization. }
  \vspace{-30pt}
  \label{fig:alpha}
\end{wrapfigure}
\paragraph{\textbf{Alpha Regularization.}}
We ablate the alpha regularization used during training.
Without alpha regularization, floaters are observable around the objects.
The model can successfully remove those floaters with the help of alpha regularization as shown in \cref{fig:alpha}, leading to a improvement over all metrics (\cref{tab:abla}, left, rows 4--5).

\paragraph{\textbf{Upsampler Architecture.}}
There is an alternative design of the upsampler, which mimics the conventional 2D upsamplers by replacing the transformers with CNNs.
We find that CNN-based upsampler leads to worse results (\cref{tab:abla}, right).
We conjecture that the transformer can capture multi-view correspondences and further enhance the spatial details.

\paragraph{\textbf{Depth \textit{vs.} XYZ.}}
In \cref{tab:abla} (right), we conduct an ablation where we predict the 3D coordinates of each Gaussian instead of the depth value.
We observe a performance drop across all metrics.
Without the constraint of camera rays, the positions of the Gaussians in 3D space become unstructured, making it prone to getting stuck at local minima,  resulting in a degraded performance.

\section{Discussion}
\label{sec:con}
In this paper, we introduce the Gaussian Reconstruction Model (\method{})---a new feed-forward 3D generative model that achieves state-of-the-art quality and speed. 
At the core of \method{} is a sparse-view 3D reconstructor, which leverages a novel transformer-based architecture to reconstruct 3D objects represented by pixel-aligned Gaussians.
We plan to release the code and trained models to make this advancement in 3D content creation available to the community.

\paragraph{\textbf{Limitations and Future Work.}} 
The output quality of our sparse-view reconstructor suffers when the input views are inconsistent. The reconstructor is deterministic in nature and future work could embed it in a probabilistic framework, akin to DMV3D~\cite{xu2023dmv3d}. Our current framework is limited to object-centric scenes due to the lack of large-scale 3D scene datasets. Future work could explore the generation of larger and more complicated scenes.

\paragraph{\textbf{Ethics.}} Generative models pose a societal threat---we do not condone using our work to generate deep fakes intending to spread misinformation.

\paragraph{\textbf{Conclusion.}} 
Our work represents a significant advancement towards efficient and high-quality 3D reconstruction and generation.

\paragraph{\textbf{Acknowledgement.}} 
We would like to thank Shangzhan Zhang for his help with the demo video, and Minghua Liu for assisting with the evaluation of One-2-3-45++. This project was supported by Google, Samsung, and a Swiss Postdoc Mobility fellowship.

\title{GRM: Large Gaussian Reconstruction Model for \texorpdfstring{\\}~Efficient 3D Reconstruction and Generation Supplementary Material}
\titlerunning{\method}

\author{
    Yinghao Xu$^{1}\thanks{Equal Contribution}$ \and
    Zifan Shi$^{1,2\star}$ \and
    Yifan Wang$^{1}$ \and Hansheng Chen$^{1}$ \\ Ceyuan Yang$^{3}$ \and Sida Peng$^{4}$ \and Yujun Shen$^{5}$ \and Gordon Wetzstein$^{1}$
}

\authorrunning{Yinghao Xu et al.}

\institute{
    Stanford University \and 
    The Hong Kong University of Science and Technology \and
    Shanghai AI Laboratory \and 
    Zhejiang University \and
    Ant Group
}

\maketitle

\appendix

This supplementary material is organized as follows.
We first introduce implementation details of our \method{} (\cref{sec:implementation}) . 
Then, we evaluate the geometry quality  of our \method{} against the baselines (\cref{sec:geometry}). 
We also present the details of mesh extraction from 3D Gaussians in \cref{sec:mesh}.
Finally, we show additional results on 3D reconstruction and generation to evaluate the flexibility and effectiveness of our approach (\cref{sec:additional-results}).

\section{Implementation Details} \label{sec:implementation}

\paragraph{\textbf{Network Architecture and Training Details.}}
We illustrate the details of network architecture and training in \cref{tab:impl_details}.
\begin{table}[h]
\vspace{-10pt}
\centering
\caption{\textbf{Implementation details.} }
\begin{tabular}{l|l|l}
\hline
\multirow{2}{*}{Encoder}   & Convolution layer      & 1, kernel size 16, stride 16    \\
                           & Att layers      & 24, channel width 768, \# heads 12     \\
\hline
\multirow{4}{*}{Upsampler block}   & Pixelshuffle per block &  1, scale factor 2       \\
                           & Att layers per block      & 2, \# heads 12\\
                           & Channel width & starting from 768, decay ratio of 2 per block \\
                           & \# Blocks & 4\\
\hline
\multirow{5}{*}{Gaussian splatting}  & Color activation & sigmoid          \\
                           & Rotation activation     & normalize       \\
                           & Opacity activation      & sigmoid        \\
                           & Scale activation & sigmoid \\
                           & Position activation & None \\
\hline
\multirow{8}{*}{Training details} & Learning rate              &    3e-4    \\
                           & Learning rate scheduler  & Cosine \\
                           & Optimizer            &    AdamW       \\
                           & (Beta1, Beta2)  & (0.9, 0.95) \\
                           & Weight decay  &     0.05      \\
                           & Warm-up       &      3000       \\
                           & Batch size    & 8 per GPU \\
                           & \# GPU        & 32 \\

\hline
\end{tabular}
\label{tab:impl_details}
\vspace{-20pt}
\end{table}


\paragraph{\textbf{Deferred Backpropagation.}}

Our model generates $4\times 512 \times 512$ Gaussians, consuming a significant amount of GPU memory.
We would only be able to train our model with a batch size of 2 on 80GB A100 GPUs.
Deferred backpropagation~\cite{zhang2022arf} is an important technique for saving GPU memory in large batch-size training.
With it, we are able to scale up the batch size to 8, consuming only 38GB per GPU.
We provide a pseudo code (\cref{code}) to demonstrate how we implement it in our model training.

\begin{algorithm}[t]
\caption{Pseudocode of Deferred Backpropagation on Gaussian Rendering in PyTorch-like style.}
\label{code}

\fontsize{7.2pt}{0em}\selectfont \texttt{Render}: Rendering process; 
\definecolor{codeblue}{rgb}{0.25,0.5,0.5}
\lstset{
  backgroundcolor=\color{white},
  basicstyle=\fontsize{7.2pt}{7.2pt}\ttfamily\selectfont,
  columns=fullflexible,
  breaklines=true,
  captionpos=b,
  commentstyle=\fontsize{7.2pt}{7.2pt}\color{codeblue},
  keywordstyle=\fontsize{7.2pt}{7.2pt},
}
\begin{lstlisting}[language=python]
class DBGaussianRender(torch.autograd.Function):
    def forward(ctx, gaussians, cameras):
        # save for backpropagation
        ctx.save_for_backward(gaussians, cameras)
        
        with torch.no_grad():
            images = Render(gaussians, cameras)

        return images
        
    def backward(ctx, grad_images):
        # restore input tensor
        gaussians, cameras = ctx.saved_tensors
        
        with torch.enable_grad():
            images = Render(gaussians, cameras)
            images.backward(grad_images)
        
        return gaussians.grad
        
\end{lstlisting}
\end{algorithm}

\paragraph{\textbf{Perceptual loss.}}
We have experimented with an alternative loss to the conventional perceptual loss mentioned in the paper, known as the Learned Perceptual Image Patch Similarity (LPIPS) loss~\cite{lpips}.
However, we observe severe unstable training and the model cannot converge well.
%

\section{Geometry Evaluation}\label{sec:geometry}

Here, we demonstrate the geometry evaluation results on sparse-view reconstruction and single-image-to-3D generation. %
We report Chamfer Distance (CD) and F-score as the evaluation metrics. %
Specifically, we use different thresholds for F-score to reduce the evaluation uncertainty.
We use ICP alignment to register all the 3D shapes into the same canonical space. %
All metrics are evaluated on the original scale in the GSO dataset. %

\paragraph{\textbf{Sparse-view Reconstruction.}} We compare with SparseNeuS~\cite{sparseneus} which trained in One-2-3-45~\cite{one2345}, and LGM~\cite{tang2024lgm} in \cref{tab:geo-recon}. 
The SparseNeuS exhibits a very high CD score with 16 views for reconstruction (32 views in the original paper) because of the far-away floaters. 
GRM achieves better geometry scores across all metrics, particularly on the F-score with small thresholds.

\begin{table}[thb!]
\centering

\caption{\textbf{Geometry evaluation on Sparse-view Reconstruction}. SparseNeuS~\cite{sparseneus, one2345} exhibits an exceptionally high CD due to the far-away floaters.}
\label{tab:geo-recon}
\begin{tabular}{lcccc}
\toprule
Method              & \#Views & CD\textdownarrow & F-Score(0.01)\textuparrow  & F-Score(0.005) \textuparrow   \\ \midrule
SparseNeuS~\cite{one2345, sparseneus} & 16 & 0.02300 &  0.3674 & 0.5822  \\
LGM ~\cite{tang2024lgm}   & 4 & 0.00393 & 0.9402 & 0.7694 \\
\method{} (Ours)  & 4 & \textbf{0.00358} &  \textbf{0.9560} & \textbf{0.8239}\\
\bottomrule
\end{tabular}
\vspace{-5pt}
\end{table}

\paragraph{\textbf{Single-Image-to-3D Generation.}}

We compare \method{} against baselines on geometry quality in \cref{tab:geo-imagen}.
The original implementation of One-2-3-45++~\cite{liu2023one} suffers from a limitation where it can only generate a single component in multi-object scenes, resulting in geometry metrics that are not as good as other baseline methods.
\method{} outperforms all baseline methods across all metrics.
Moreover, when compared to optimization-based methods, such as DreamGaussian~\cite{dreamgaussian}, our approach demonstrates notable runtime advantages along with superior geometry quality.

\begin{table}[thb!]
\centering

\caption{\textbf{Geometry evaluation on Single-image-to-3D generation}. Note that the original implementation of One-2-3-45++~\cite{liu2023one} suffers from a limitation where it can only generate a single component in multi-object scenes.}
\label{tab:geo-imagen}
\begin{tabular}{lccc}
\toprule
Method              & CD\textdownarrow &F-Score(0.01)\textuparrow  & F-Score(0.025) \textuparrow   \\ \midrule
One-2-3-45++~\cite{liu2023one}           & 0.0145  & 0.6419 & 0.8362 \\
Wonder3D~\cite{long2023wonder3d}         & 0.0131 &0.6384 & 0.8576 \\
One-2-3-45~\cite{one2345}             &  0.0134 & 0.6689  & 0.8682 \\
Shap-E~\cite{shape}               & 0.0118 & 0.6990 &  0.8820 \\
\midrule
LGM ~\cite{tang2024lgm}                & 0.0123 & 0.6853 & 0.8591\\
DreamGaussian~\cite{dreamgaussian}       & 0.0077  & 0.7616 & 0.9506\\
\method{} (Ours)    & \textbf{0.0058} & \textbf{0.8758} & \textbf{0.9775}  \\

\bottomrule
\end{tabular}
\vspace{-5pt}
\end{table}

\begin{figure}[h!]
	\centering
	\includegraphics[width=0.9\linewidth]{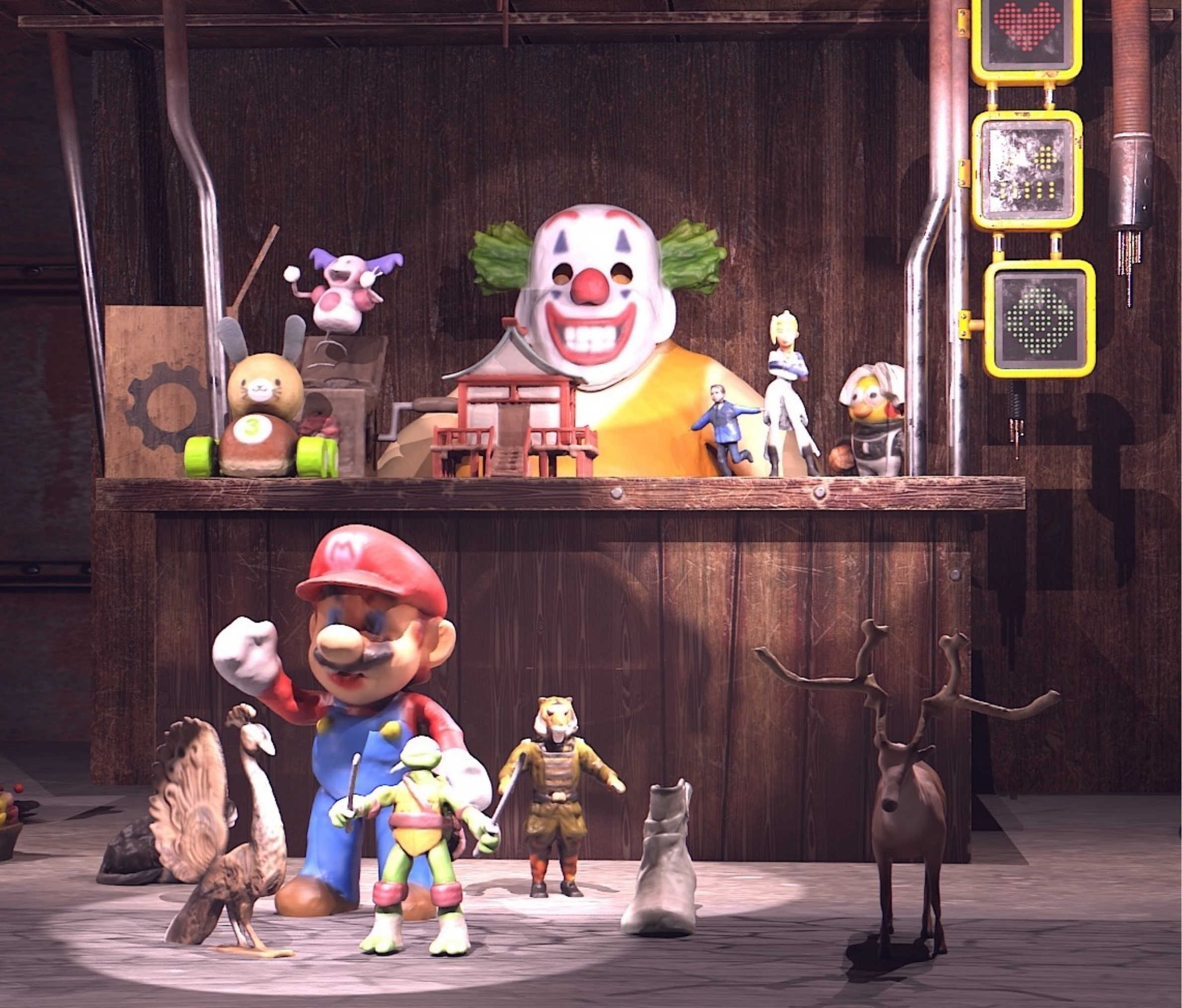}
	\caption{\textbf{Blender scene constructed with our textured mesh.}}
	\label{fig:blender}
\end{figure}

\section{Mesh Extraction from 3D Gaussians}\label{sec:mesh}

We utilize the Fibonacci sampling method to sample 200 uniformly distributed cameras on sphere for rendering images and depth maps based on the 3D Gaussians of the scene. Subsequently, we fuse the RGB-D data using the TSDFVolume~\cite{curless1996volumetric} method to generate a mesh.
We must take into account that due to the Gaussian distribution, some points may scatter outside the surface of the object. 
Therefore, we employ clustering to remove very small floaters outside the object's surface in order to smooth the generated mesh.

\section{Additional Visual Results}\label{sec:additional-results}
\begin{figure}[h!]
	\centering
	\includegraphics[width=1.0\linewidth]{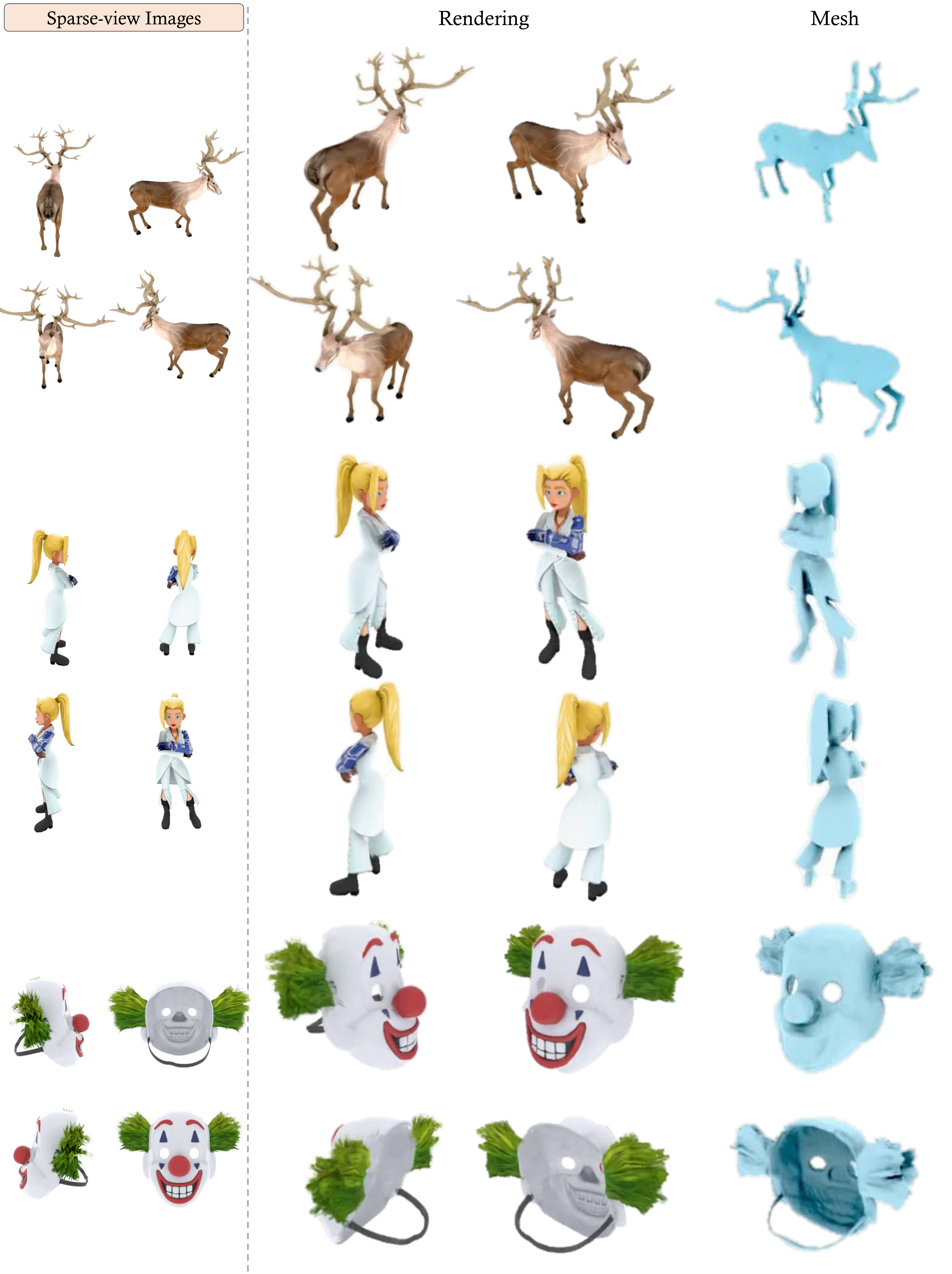}
	\caption{\textbf{Sparse-view Reconstruction.}}
	\label{fig:sparse-view reconstructor-more}
\end{figure}

\begin{figure}[h!]
	\centering
	\includegraphics[width=1\linewidth]{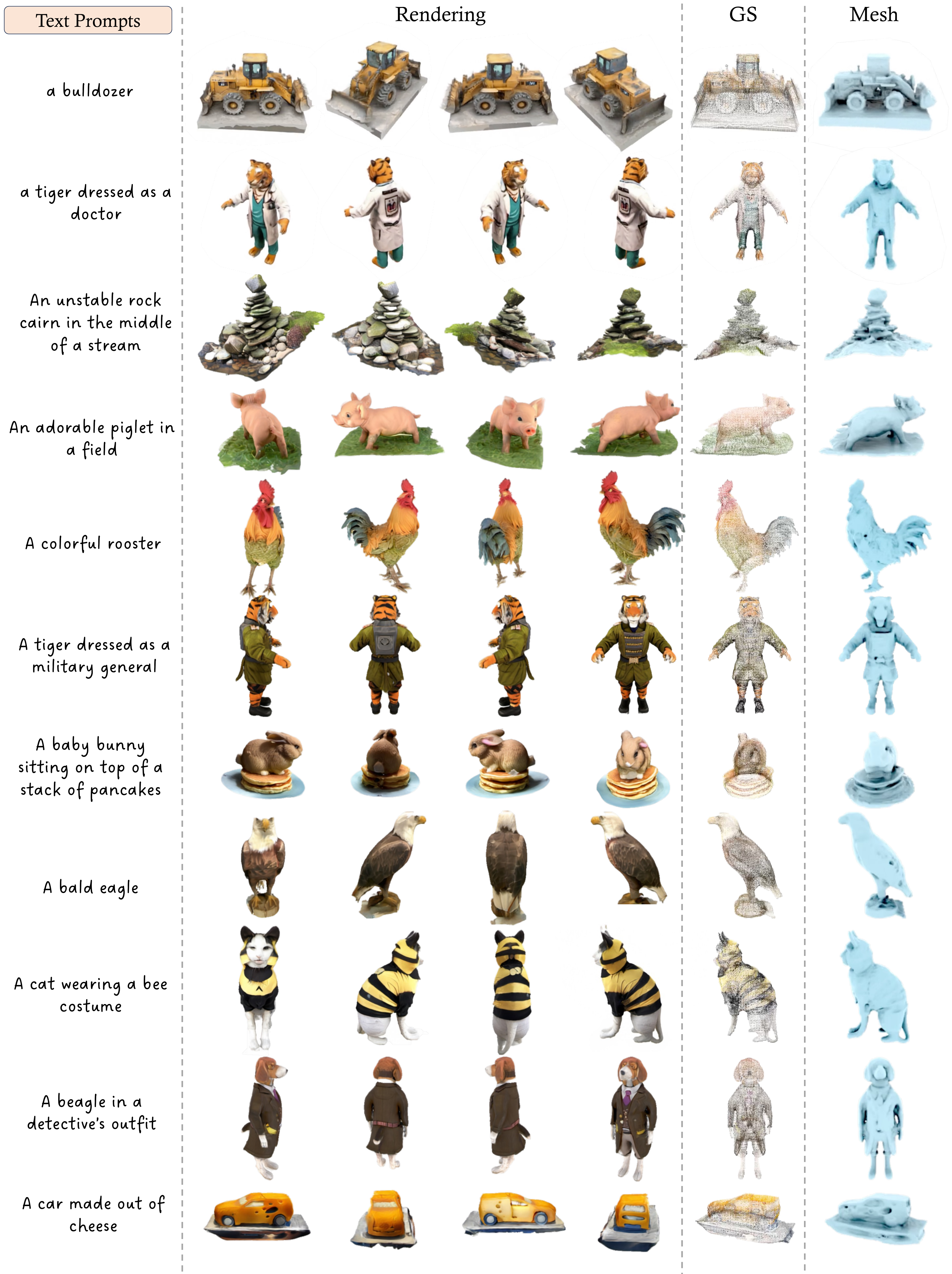}
	\caption{\textbf{Text-to-3D Generation.}}
	\label{fig:text-to-3d-more}
\end{figure}

\begin{figure}[h!]
	\centering
	\includegraphics[width=1\linewidth]{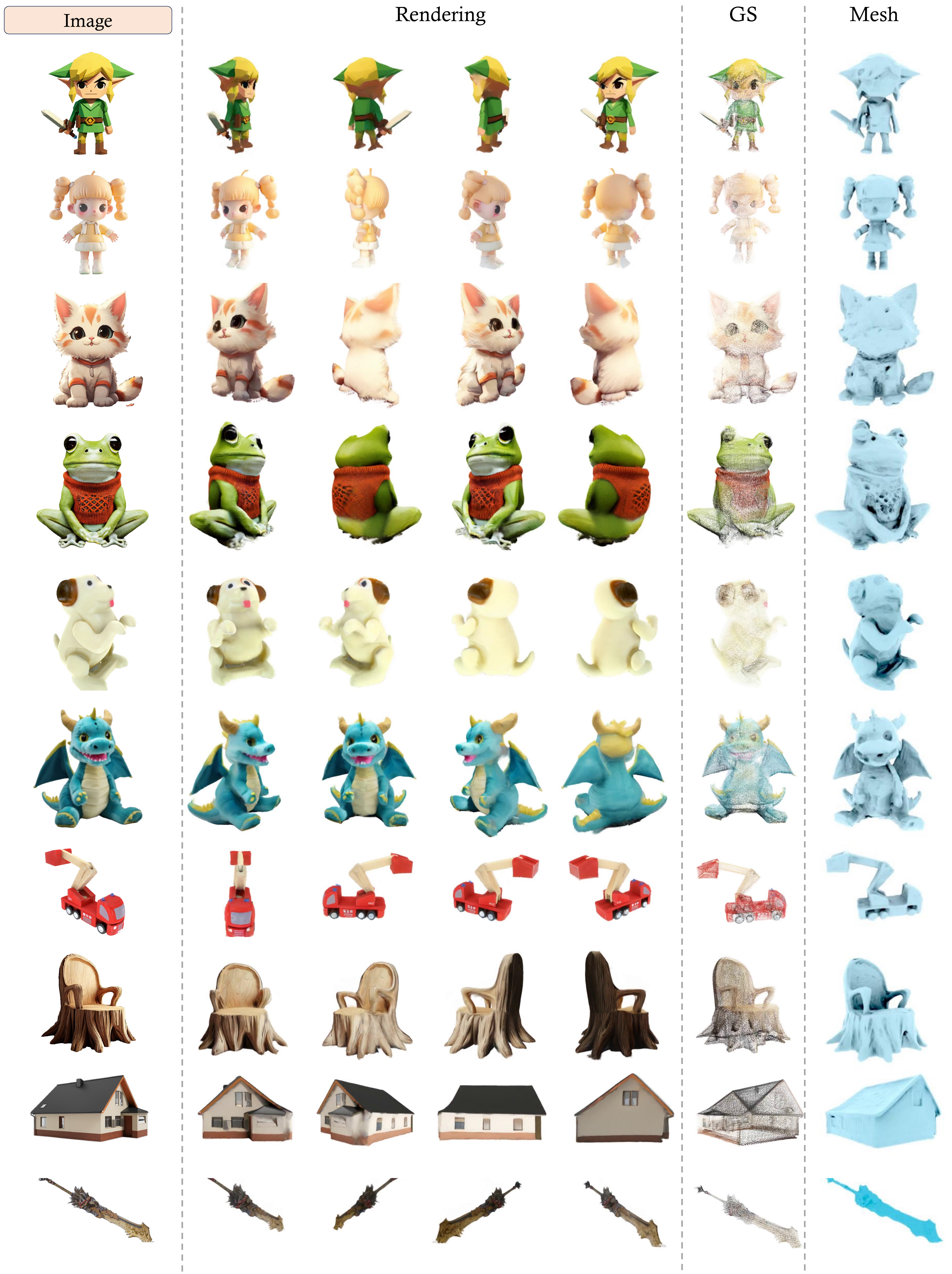}
	\caption{\textbf{Single-image-to-3D Generation.}}
	\label{fig:image-to-3d-more}
\end{figure}

We assemble the extracted texture mesh in Blender to construct a 3D scene.
We attach the scene image in \cref{fig:blender}.
We include more qualitative results on sparse-view reconstruction, text-to-3D generation and image-to-3D generation in \cref{fig:sparse-view reconstructor-more}, \cref{fig:text-to-3d-more} and \cref{fig:image-to-3d-more}, respectively.

\section{Limitations}
Despite the high-quality reconstruction, image-to-3D and text-to-3D generation results we achieved, our model relies on the input information for reconstruction and lacks the capability for hallucination.
For example, if a region is not observed in any of the input images, the model may produce blurry textures for it.

\bibliographystyle{splncs04}
\bibliography{ref.bib}

\end{document}